\documentclass[letterpaper, 10 pt, conference]{ieeeconf}  % Comment this line out if you need a4paper
\usepackage{mathtools}
\usepackage{ifluatex,ifxetex}
\ifluatex
  \usepackage{fontspec} % optional
\else\ifxetex
  \usepackage{fontspec} % optional
\else % assume that pdftex engine is in use
  \usepackage[T1]{fontenc}
  \usepackage[utf8]{inputenc} % optional for "\ng"
\fi\fi

\IEEEoverridecommandlockouts                              % This command is only needed if 
\usepackage{amssymb}
\usepackage{color}
\let\labelindent\relax
\usepackage[inline]{enumitem}

\title{\LARGE \bf
Grouptron: Dynamic Multi-Scale Graph Convolutional Networks for Group-Aware Dense Crowd Trajectory Forecasting
}

\author{Rui Zhou$^{1}$, Hongyu Zhou$^{2}$, Huidong Gao$^{3}$, Masayoshi Tomizuka$^{3}$, Jiachen Li$^{4,*}$, and Zhuo Xu$^{3,*}$% <-this % stops a space
\thanks{$^{*}$These authors contributed equally to this paper}%
\thanks{$^{1}$Rui Zhou is with the Department of Electrical Engineering and Computer Science, University of California, Berkeley, CA 94720 USA \tt\small {ruizhouzr@berkeley.edu}}

\thanks{$^{2}$Hongyu Zhou is with the Department of Aerospace Engineering, University of Michigan, Ann Arbor, MI 48109 USA
        {\tt\small zhouhy@umich.edu}}%
        
\thanks{$^{3}$Huidong Gao, Masayoshi Tomizuka, and Zhuo Xu are with the Department of Mechanical Engineering, University of California, Berkeley, CA
94720 USA 
        {\tt\small \{hgao9, tomizuka, zhuoxu\}@berkeley.edu}}
        
\thanks{$^{4}$Jiachen Li is with the Department of Aeronautics \& Astronautics, Stanford University, CA 94305 USA 
        {\tt\small jiachen\_li@stanford.edu}}

}

\begin{document}

\maketitle
\thispagestyle{empty}
\pagestyle{empty}

%%%%%%%%%%%%%%%%%%%%%%%%%%%%%%%%%%%%%%%%%%%%%%%%%%%%%%%%%%%%%%%%%%%%%%%%%%%%%%%%
\begin{abstract}
Accurate, long-term forecasting of pedestrian trajectories in highly dynamic and interactive scenes is a long-standing challenge. Recent advances in using data-driven approaches have achieved significant improvements in terms of prediction accuracy. However, the lack of group-aware analysis has limited the performance of forecasting models. This is especially nonnegligible in highly crowded scenes, where pedestrians are moving in groups and the interactions between groups are extremely complex and dynamic. 
In this paper, we present Grouptron, a multi-scale dynamic forecasting framework that leverages pedestrian group detection and utilizes individual-level, group-level and scene-level information for better understanding and representation of the scenes. 
Our approach employs spatio-temporal clustering algorithms to identify pedestrian groups, creates spatio-temporal graphs at the individual, group, and scene levels. 
It then uses graph neural networks to encode dynamics at different scales and aggregate the embeddings for trajectory prediction. 
We conducted extensive comparisons and ablation experiments to demonstrate the effectiveness of our approach. Our method achieves 9.3\% decrease in final displacement error (FDE) compared with state-of-the-art methods on ETH/UCY benchmark datasets, and 16.1\% decrease in FDE in more crowded scenes where extensive human group interactions are more frequently present.

\end{abstract}

%%%%%%%%%%%%%%%%%%%%%%%%%%%%%%%%%%%%%%%%%%%%%%%%%%%%%%%%%%%%%%%%%%%%%%%%%%%%%%%%
\section{Introduction}

Generating long-term and accurate predictions of human pedestrian trajectories is of enormous significance for developing reliable autonomous systems (e.g. autonomous vehicles, mobile robots), which are often required to safely navigate in crowded scenarios with many human pedestrians present, such as in crowded open traffic or in warehouses. Therefore, the capability of understanding and predicting the dense crowd behavior is instrumental in order to avoid collisions with the highly dynamic crowds and to behave in a socially-aware way.

Through daily interactions, humans are able to develop the remarkable capability of understanding crowded and interactive environments and inferring the potential future movements of all other traffic participants. 
Similarly, recent advances using machine learning techniques leverage the massive human interaction data are capable of achieving state-of-the-art performance on trajectory prediction \cite{alahi2016social, salzmann2020trajectron++,li2020evolvegraph,li2021rain,choi2021shared,ma2021continual}. 
However, most state-of-the-art methods fail to consider the densely populated scenarios, which are extremely challenging due to the immensely dynamic and complex interactions. Moreover, in such highly crowded scenarios, groups of pedestrians are common. \cite{Sochman2011WhoKW} estimates that 50\% to 70\% of pedestrians walk in groups, which exhibit dramatically distinct behaviors from individuals. 
However, existing methods lack the mechanism of representing such group dynamics, which limits their performance in highly crowded situations.

\begin{figure}[t]
\centering
\includegraphics[width=\columnwidth]{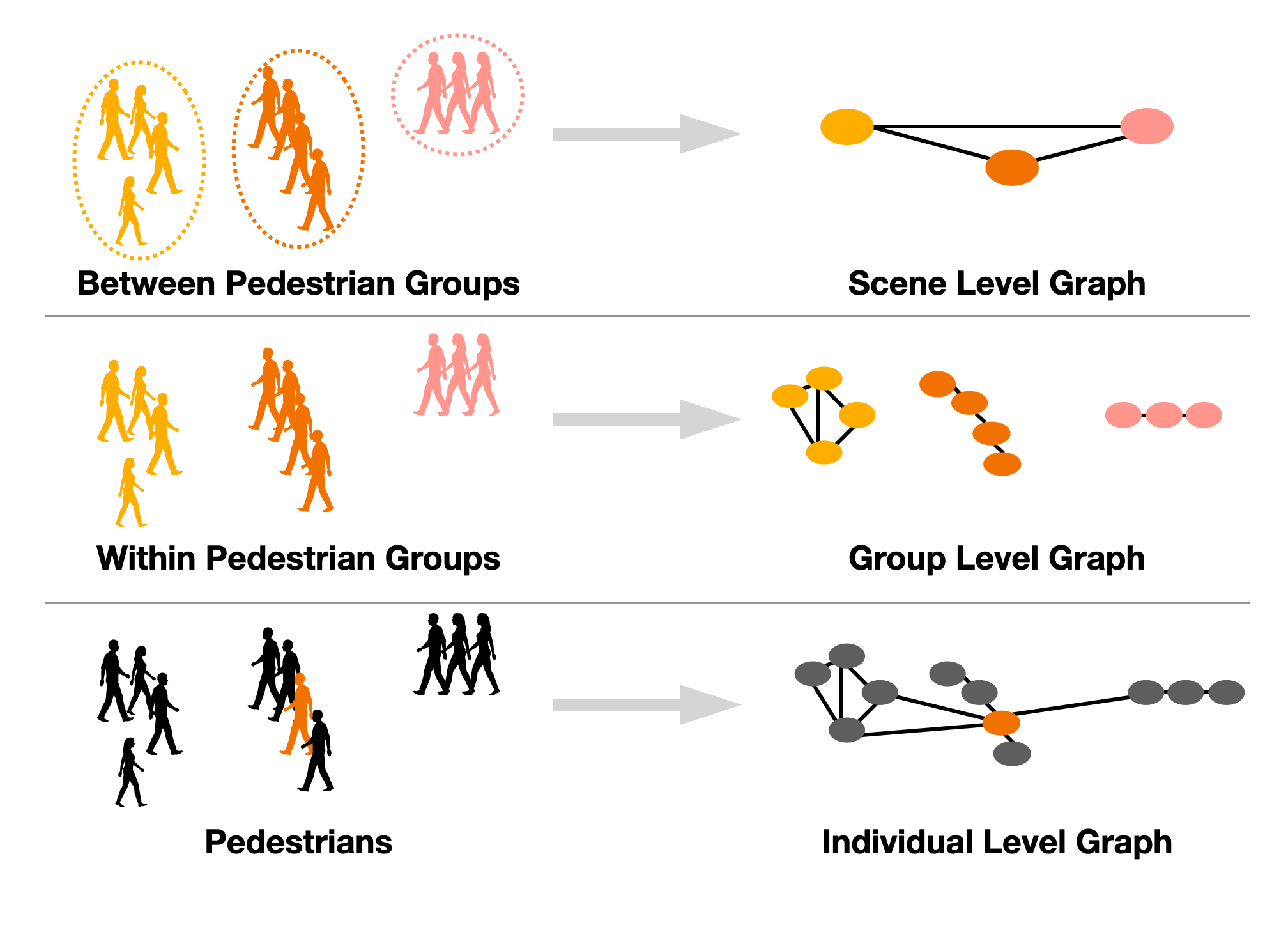}
\caption{Grouptron models the pedestrian behaviors in a multi-scale fashion and constructs spatio-temporal graphs for different scales.}\label{fig:model}
\end{figure}

In this paper, we seek to resolve this challenge by leveraging group detection and dynamic spatio-temporal graph representations at different scales as visualized in Fig. \ref{fig:model}.
Concretely, we construct a dynamic multi-scale graph convolutional neural network that uses pedestrian group information to better learn and model the extremely complex dynamics at different scales and between different scales. 

The contributions of this paper are summarized as follows:
\begin{itemize}
    \item We propose to leverage spatio-temporal clustering algorithms to detect pedestrians in groups. 
    \item We design a hierarchical prediction framework using spatio-temporal graph neural networks which encode the scene at three levels: 
        \begin{enumerate}
            \item The \textit{individual} level encodes the historical trajectory of the predicted pedestrian.
            \item The \textit{group} level encodes the dynamics and trajectory information within each pedestrian group.
            \item The \textit{scene} level encodes the dynamics and interactions between pedestrian groups.
        \end{enumerate}
    \item The proposed multi-scale spatio-temporal architecture outperforms existing methods by 9.3\% in terms of final displacement errors. In particular, for densely crowded scenarios, the performance improvement can be as significant as 16.1\%.
\end{itemize}

\section{Related Work}
   \begin{figure*}[thpb]
      \centering
      \includegraphics[width=0.827\textwidth]{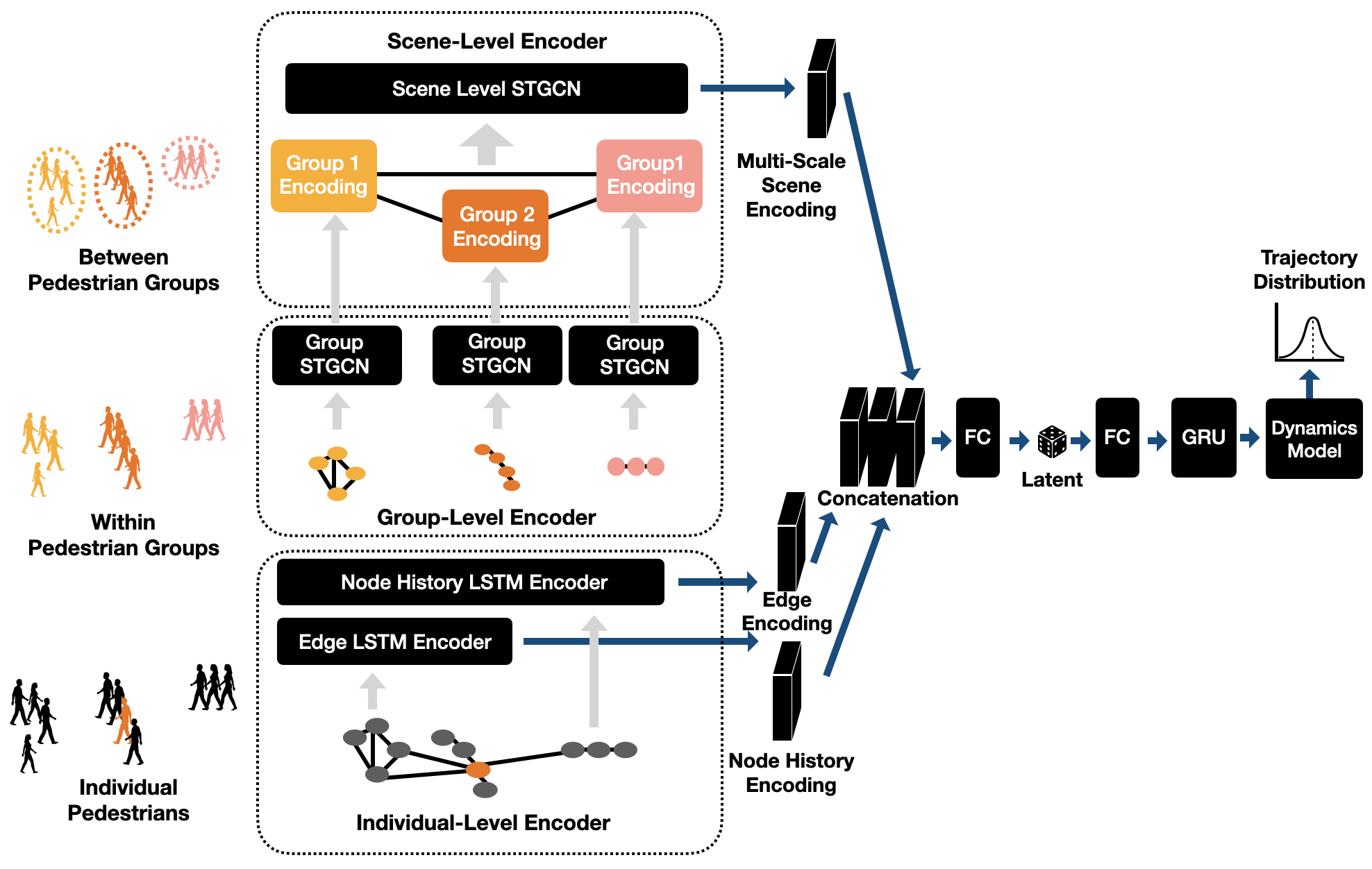}
      \caption{The diagram of the Grouptron model, which creates spatio-temporal graphs for different levels and encodes these graphs using different networks. Combining encodings for the individual level, the group level, and the scene level, Grouptron then uses a GRU decoder to output the predictions.}
      \label{network}
   \vspace{-4mm}
   \end{figure*}
   
\subsection{Human Trajectory Forecasting}
One of the pioneering works of human trajectory forecasting is the Social Force model \cite{helbing1995social}, which applies Newtonian forces to model human motion. Similar methods with strong priors have also been proposed \cite{antonini2006discrete}; yet, most of them rely on hand-crafted energy potential, such as relative distances and rules, to model human motion.

Recently, machine learning methods have been applied to the problem of human trajectory forecasting to obtain models with better performance. 
One line of work is to formulate this problem as a deterministic time-series regression problem and then solve it using, e.g., Gaussian Process Regression (GPR) \cite{wang2007gaussian}, inverse reinforcement learning (IRL) \cite{lee2016predicting}, and recurrent neural networks (RNNs) \cite{alahi2016social,jain2016structural,vemula2018social}.

However, the issue of these deterministic regressors is that human behavior is rarely deterministic or unimodal. Hence, generative approaches have become the state-of-the-art trajectory forecasting methods, due to recent advancements in deep generative models \cite{sohn2015learning, goodfellow2014generative} and their ability of generating distributions of potential future trajectories (instead of a single future trajectory). Most of these methods use a recurrent neural network architecture with a latent variable model, such as a conditional variational auto-encoder (CVAE) \cite{salzmann2020trajectron++,lee2017desire,ma2019wasserstein,li2021spatio}, or a generative adversarial network (GAN) \cite{gupta2018social,kosaraju2019social,sadeghian2019sophie,zhao2019multi,li2019conditional} to encode multi-modality. Compared to previous work, we not only consider multi-modality from the perspective of a single agent, but also from the group level; we take into account the phenomenon that people usually move in groups. We show that our group-aware prediction has better understanding of the scenes and achieves better forecasting performance.

\subsection{Graph Convolutional Networks}
Of the methods mentioned above, RNN-based methods have achieved better performance. However, recurrent architectures are parameter inefficient and expensive in training \cite{bai2018empirical}. Besides, to handle spatial context, RNN-based methods need additional structures. Most of them use graph models to encode neighboring pedestrians' information since the topology of graphs is a natural way to represent interactions between pedestrians. 
Graph convolutional networks (GCN) introduced in \cite{kipf2016semi} is more suitable for dealing with non-Euclidean data. The Social-BiGAT \cite{kosaraju2019social} introduces a graph attention network \cite{velivckovic2017graph} to model social interactions. GraphSAGE \cite{hamilton2017inductive} aggregates nodes and fuses adjacent nodes in different orders to extract node embeddings. To capture both the spatial and temporal information, Spatio-Temporal Graph Convolutional Networks (STGCN) extends the spatial
GCN to spatio-temporal GCN for skeleton-based action recognition \cite{yan2018spatial}. STGCN is adapted by Social-STGNN \cite{mohamed2020social} for trajectory forecasting, where trajectories are modeled by graphs with edges representing social interactions and weighted by the distances between pedestrians. A development related to our paper is dynamic multi-scale GNN (DMGNN) \cite{li2020dynamic}, which proposes a \textit{multi-scale graph} to model human body relations and extract features at multiple scales for motion prediction. There are two kinds of sub-graphs in the multi-scale graph: 
\begin{enumerate*}[label=(\roman*)]
\item single-scale graphs, which connect body components at the same scales, and
\item cross-scale graphs, which form cross-scale connections among body components.
\end{enumerate*}
Based on the multi-scale graphs, a multi-scale graph computational unit is proposed to extract and fuse features across multiple scales. Motivated by this work, we adopt the multi-scale graph strategy for dense crowd forecasting which includes scene-level graphs, group-level graphs, and individual-level graphs.

\subsection{Group-aware Prediction}
People moving in groups (such as friends, family members, etc.) is a common phenomenon and people in each group tend to exhibit similar motion patterns. Motivated by this phenomenon, group-aware methods \cite{rudenko2020human} consider the possibility of human agents being in groups or formations to have more correlated motions than independent ones. They therefore can also model reactions of agents to the moving groups. Human agents can be assigned to different groups by clustering trajectories with similar motion patterns based on methods such as $k$-means clustering \cite{zhong2015learning}, support vector clustering \cite{lawal2016support}, coherent filtering \cite{bisagno2018group}, and spectral clustering methods \cite{atev2010clustering}.

\section{Grouptron}
\subsection{Problem Formulation}
We aim to predict the trajectory distributions of a time-varying $N(t)$ number of pedestrians $P_{N(t)}$. At time step $t$, given the 2-D position $\mathbf{s} \in \mathbb{R}^2$ of each pedestrian and their previous trajectories of $T$ time steps $\mathbf{T} \in \mathbb{R}^{T \times 2}$, our goal is to predict the distributions of their trajectories in the next $F$ future time steps, $y=\mathbf{T}_{1,..,N(t)}^{t:t+F} \in \mathbb{R}^{N(t)\times F \times 2}$. We denote the distributions as $p(y\mid x)$. The performance of the predictions is evaluated by standard distance-based metrics on benchmark datasets: the average displacement error (ADE) and the final displacement error (FDE).
\subsection{Model Overview}
Rooted in the CVAE architecture in Trajectron++  \cite{salzmann2020trajectron++}, we design a more expressive multi-scale scene encoding structure, which actively takes into consideration the group-level and scene-level information for better representation of crowded scenes where groups of pedestrians are present. 
Concretely, we leverage spatio-temporal graphs for each level to model information and interactions at the corresponding level. We refer to our model as Grouptron. Our model is illustrated in Fig. \ref{network}. In this subsection, we provide an overview of the architecture, and in 
Section III-C, we elaborate on the details of the Grouptron model. 

At the individual level, we construct spatio-temporal graphs for individual pedestrians. The graph is centered at the node whose trajectory we want to predict. We call it the ``current node''. Long short term memory (LSTM) networks \cite{hochreiter1997long} are used to encode this graph. We group the pedestrians with the agglomerative clustering algorithm based on Hausdorff distances \cite{atev2010clustering}. STGCN is used to encode dynamics within the groups.

At the scene level, spatio-temporal graphs are created to model dynamics among pedestrian groups and are encoded using a different STGCN. Lastly, the information across different scales is combined. A decoder is then used to obtain trajectory predictions and the model can output the most possible trajectory or the predicted trajectory distributions. 

\subsection{Multi-Scale Scene Encoder}
\subsubsection{Individual-Level Encoder}
The first level of encoding is for the individual pedestrians. We represent information at the individual level using a spatio-temporal graph for the current node. The nodes include the current node and all other nodes that are in the perception range of the current node and nodes whose perception range covers the current node. The node states are the trajectories of the nodes. The edges are directional and there is an edge $e_{i,j}$ if pedestrian $i$ is in the perception range of pedestrian $j$.
To encode the current node's history trajectory, we use an LSTM network with hidden dimension 32. To encode the edges, we first perform an element-wise sum on the states of all neighboring nodes to form a single vector that represents all neighbors of the current node. This vector is then fed into the edge LSTM network which is an LSTM network with hidden dimension 8.
In this way, we obtain two vectors: a vector encoding the trajectory history of the current node and a vector encoding the representation of all the neighbors of the current node.
\subsubsection{Pedestrian Group Clustering}
To cluster nodes into groups based on trajectories, we propose to leverage the agglomerative clustering algorithm \cite{atev2010clustering}, which uses similarity scores based on Hausdorff distances between trajectories. The number of clusters (groups) to create for each scene is determined by:
\begin{equation}
    C(N) = (N+1)/2,
\end{equation}
where C is the number of clusters and N is the total number of nodes to be clustered.

Furthermore, We only include nodes with an edge to or from the current node. This is because we only want to include nodes that can potentially influence the current node to avoid unhelpful information from nodes that are too far away from the current node.

\subsubsection{Group-Level Encoder}
For each group, we create a spatio-temporal graph consisting of $G_{g, t} = (V_{g,t}, E_{g, t})$, where $t$ is the time step and $g$ is the group id. $V_{g,t}= \{v_{i,t} \mid \forall i \in \{1,...,N_{g}\}\}$ are all the nodes in the group $g$. The node states are the trajectories of the represented pedestrians. $E_{g, t}=\{e^{i,j}_t \mid \forall i,j \in \{1,...,N_{g}\}$ are the set of edges between the nodes in the current group such that $e^{i,j}_t = 1$ to allow maximum interaction modeling within pedestrian groups.

After forming the aforementioned graphs for each group, they are then passed to the group-level trajectory encoder to obtain the encoded vectors for nodes in each group. The group-level trajectory encoder is an STGCN proposed in \cite{yan2018spatial} and used in \cite{mohamed2020social}. We set the convolution filter size to 3 and use the same weight parameters for all the groups.

We then average the encoded vectors of all nodes in each group to obtain the representations for the corresponding groups. That is, $E_{g} = \frac{1}{N_{g}} \sum_{i=1}^{N_{g}} E_{i}$, where $E_{g}$ is the encoded vector for group $g$, $E_{i}$ is the encoded vector for node $i$ in the output from the group-level trajectory encoder, and $N_g$ is the number of nodes in group $g$.

\subsubsection{Scene-Level Encoder}
After obtaining the encoded vectors for each group, a scene-level spatio-temporal graph with nodes representing groups is created. That is, $G_{scene, t} = (V_{scene,t}, E_{scene, t})$. $V_{scene,t}= \{v_{g, t} \mid \forall\,\text{g} \in \{1,...,G\}\}$, where G is the total number of groups and $t$ is the timestep. The state for each node is $E_{g}$ from the group-level trajectory encoder. $E_{scene, t}=\{e^{i,j}_t \mid \forall i,j \in \{1,...,G\}$ are the set of edges between the groups in the scene. Each $e^{i,j}$ is set to $1$ to allow maximum message passing between group nodes.

We then select the encoded vector corresponding to the last timestep and the group id of the current node as the scene-level encoding: $E_{scene} = E_{g, T}$, where $g$ is the group id of the current node we are encoding for and T is the total number of time steps. 

\subsubsection{Multi-Scale Encoder Output}
The output of the multi-scale scene encoder is the concatenation of the following level encoded vectors: the output from the node history encoder, the output from individual-level edge encoder, and the output from the scene-Level encoder. That is, $    E_{multi} = [E_{his};E_{edge};E_{scene}]$, where $E_{his}$ is the encoded vector for the current node's history trajectory, $E_{edge}$ is the vector representing individual-level neighbors, $E_{scene}$ is the encoded vector from the scene-level encoder. 
\subsection{Decoder Network}
Together with the latent variable $z$, $E_{multi}$ is passed to the decoder that is a Gated Recurrent Unit (GRU) \cite{chung2014empirical} with 128 dimensions. The output from the GRU decoder is then fed into dynamics integration modules as control actions to output the predicted trajectory distributions, or the single most-likley trajectory, depending on the task.
\subsection{Loss Functions}
We adopt the following objective function for the overall CVAE model:
\begin{equation}
\begin{aligned}
    \max_{\phi,\theta,\psi} \sum_{i=1}^{N} & \mathbb{E}_{z \sim q_\phi(\cdot | \mathbf{x}_i,\mathbf{y}_i)} \left[\log p_\psi(\mathbf{y}_i | \mathbf{x}_i,z)\right] \\
    & - \beta D_{KL} \left(q_\phi\left(z | \mathbf{x}_i,\mathbf{y}_i\right)||p_\theta\left(z|\mathbf{x}_i\right)\right)+\alpha I_q(\mathbf{x}|z),    
\end{aligned}
\end{equation}
where $I_q$ is the mutual information between x and z under the distribution $q_\phi(\mathbf{x} | z)$. We follow the process given in \cite{zhao2019infovae} to compute $I_q$. We approximate $q_\phi\left(z | \mathbf{x}_i,\mathbf{y}_i\right)$ with $p_\theta\left(z|\mathbf{x}_i\right)$, and obtain the unconditioned latent distribution by summing out $\mathbf{x}_i$ over the batch.

\section{Experiments}
\begin{table*}[!htbp]
\small
\caption{FDE/ADE {\upshape (m)} values for baseline methods and Grouptron on ETH and UCY datasets. Lower is better.  Bold indicates best.}
\vspace{-0.3cm}
\label{table1}
\begin{center}
\begin{tabular}{|c||c|c|c|c|c|c|}
\hline
Method & ETH & HOTEL & UNIV & ZARA1 & ZARA2 & AVG\\
\hline
Social-LSTM&2.35/1.09 &1.76/0.79 & 1.40/0.67&1.00/0.47 & 1.17/0.56 & 1.54/0.72\\
\hline
Social-GAN&1.52/0.81 &1.61/0.72 & 1.26/0.60&0.69/0.34 & 1.84/0.42 & 1.18/0.58\\
\hline
SoPhie&\textbf{1.43}/\textbf{0.70} &1.67/0.76 & 1.24/0.54&\textbf{0.63}/0.34 & 0.78/0.38 & 1.15/0.54\\
\hline
Trajectron++&1.68/0.71 &\textbf{0.46}/0.22 &1.07/0.41 & 0.77/\textbf{0.30}& 0.59/0.23&0.91/0.37\\
\hline
Grouptron& 1.56/\textbf{0.70} & \textbf{0.46/0.21} &\textbf{0.97/0.38}& 0.76/\textbf{0.30} &\textbf{0.56/0.22}& \textbf{0.86/0.36}\\

\hline
\end{tabular}
\vspace{-0.3cm}
\end{center}
\end{table*}

\begin{table*}[!ht]
\small
\caption{FDE/ADE {\upshape (m)} values for baseline methods and Grouptron on UNIV-N Datasets. Lower is better. Bold indicates best.}
\vspace{-0.3cm}
\label{table2}
\begin{center}
\begin{tabular}{|c||c|c|c|c|}
\hline
Method & UNIV & UNIV-40 &UNIV-45& UNIV-50\\
\hline
Trajectron++ &1.07/0.41 & 1.17/0.436& 1.24/0.46&1.25/0.47\\
\hline
Grouptron &\textbf{0.97/0.38} & \textbf{1.00/0.39} &\textbf{1.04/0.40}&\textbf{1.07/0.42}\\
\hline
\end{tabular}
\vspace{-0.3cm}
\end{center}
\end{table*}

\subsection{Datasets}
The model is trained on two publicly available datasets that are benchmarks in the field: The ETH \cite{pellegrini2009you}, with subsets named ETH and HOTEL, and the UCY \cite{lerner2007crowds} datasets, with subsets named ZARA1, ZARA2, and UNIV. The trajectories are sampled at $0.4$ seconds intervals. The model observes 8 time steps, which corresponds to 3.2 seconds, and predicts the next 12 time steps, which corresponds to 4.8 seconds. 

To further evaluate and demonstrate Grouptron's performance in densely populated scenarios, we create UNIV-N test sets, where $N$ is the minimum number of people present simultaneously at each time step in the test sets. Each UNIV-N test set contains all time steps that have at least $N$ people in the scene simultaneously from the original UNIV test set. In this way, we created test sets UNIV-40, UNIV-45, and UNIV-50 and, at each time step, there are at least 40, 45, and 50 people in the scene at the same time. These test sets are far more challenging than the original UNIV test set because of the more complex and dynamic interactions at different scales. We train the models on the original UNIV training set and evaluate the models on the UNIV-40, UNIV-45, and UNIV-50 test sets.
\subsection{Evaluation Metrics}
As in prior work \cite{alahi2016social,salzmann2020trajectron++,ivanovic2019trajectron} and more, we used the following metrics to evaluate our model:
\subsubsection{Final Displacement Error}
\begin{equation}
    FDE = \frac{\sum_{i\in N}|| \hat{T^i_{T_F}}-T^i_{T_F} ||_2}{N},
\end{equation}
which is the $\textit{l}_2$ distance between the predicted final position and the ground truth final position with prediction horizon $T_F$.

\subsubsection{Average Displacement Error}
\begin{equation}
ADE = \frac{\sum_{i\in N}\sum_{t\in T_F}|| \hat{T^i_t}-T^i_t ||_2}{N\times T_F},
\end{equation}
where  $N$ is the total number of pedestrians, $T_F$ is the number of future timesteps we want to predict for, $\hat{T^i_{t}}$ is the predicted trajectory for pedestrian $i$ at timestep $t$. $ADE$ is the mean $\textit{l}_2$ distance between the ground truth and predicted trajectories.

\subsection{Experiment settings}
The Grouptron model is implemented using PyTorch.  The model is trained using an Intel I7 CPU and NVIDIA GTX 1080 Ti GPUs for 100 epochs. The batch size is 256. For the HOTEL, UNIV, ZARA1, and ZARA2 datasets, the output dimension for the group-level and scene level encoders are 16. For the ETH dataset, we set the output dimension of the group-level and scene-level encoders to be 8. This is because the ETH test set contains only 2 timesteps with at least 5 people in the scene, out of the total 1161 timesteps. In comparison, the training set contains 1910 timesteps, out of 4976 in total, with at least 5 people. Thus, to help the model learn generalizable representations in this case, we decrease the output dimension of the STGCNs to 8. The learning rate is set to $0.001$ initially and decayed exponentially every epoch with a decay rate of 0.9999. The model is trained using Adam gradient descent and gradients are clipped at $1.0$.
\subsection{Evaluation of the Group Clustering Algorithm}
\begin{figure}[t]
      \centering
      \includegraphics[width=0.75\columnwidth]{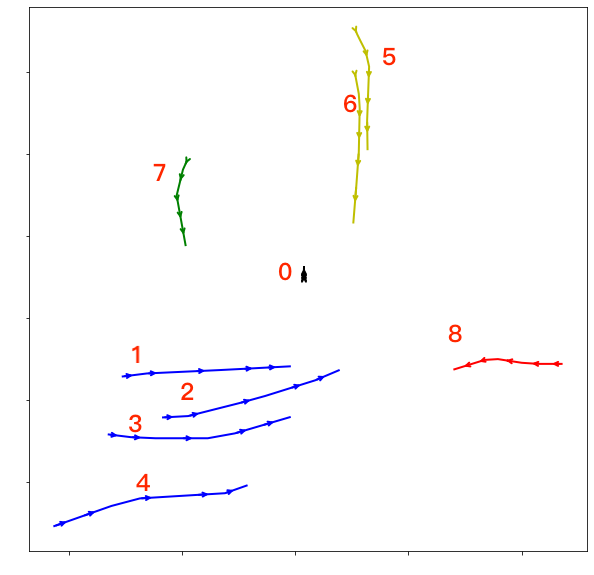}
      \caption{An example of pedestrian group clustering. Pedestrians are divided into five groups. Different colors indicate different groups.}
      \label{group}
      \vspace{-3mm}
\end{figure}
In Fig. \ref{group}, it is shown that the groups created by the agglomerative clustering method are very close to the natural definition of pedestrian groups. We can see that pedestrians 5 and 6 are travelling in a highly correlated fashion and pedestrians 1, 2, 3, and 4's trajectories are highly similar as well. In both cases, the clustering algorithm is able to correctly cluster these pedestrians into their corresponding groups. This shows that by using agglomerative clustering based on Hausdorff distances, Grouptron is able to successfully generate naturally defined groups. 
To quantitatively evaluate the groups generated by the algorithm, we selected 10 random time steps from the ETH training dataset. We invited 10 human volunteers to label groups for the time steps and used the agglomerative clustering method described in Section III-C to generate group clusters, respectively. For both human-generated groups and algorithm-generated groups, the number of groups to be formed is computed using Equation 1. We then compute the average Sørensen–Dice coefficient between human-generated and algorithm-generated groups. That is, we use
\begin{equation}
DSC = \frac{1}{T}\sum_{t\in T}\frac{1}{H}\sum_{h\in H}\frac{2\times |G_{h,t}\cap G_{a,t}|}{|G_{h,t}|+|G_{a,t}|},
\end{equation}
where $T$ is the number of timesteps, $H$ is the total number of human annotators, $G_{a, t}$ is the grouping created by the agglomerative clustering method for time step $t$, $G_{h,t}$ is the grouping created by humans for time step $t$, and $|G_{h}\cap G_{a}|$ measures how many of the groups by humans and the algorithm are exactly the same.
The Average Dice coefficient between human annotators and the algorithm is 0.72. The higher the Dice coefficient, the more similar are the groups created by the agglomerative clustering method and human annotators. Thus, The average Dice Coefficient of 0.72 indicates the groups output by agglomerative clustering method are really similar to human-generated ones.

\subsection{Quantitative Results}
We compare Grouptron's performance with state-of-the-art methods and common baselines in the field in terms of the FDE and ADE metrics, and the results are shown in Table \ref{table1}. Overall, Grouptron outperforms all state-of-the-art methods with considerable decrease in displacement errors. Since Grouptron is built on Trajectron++, we also compare the FDE and ADE values of Grouptron and Trajectron++. We find that Grouptron outperforms Trajectron++ on all 5 datasets by considerable margins. Particularly, on the ETH dataset, Grouptron achieves an FDE of 1.56m, which is 7.1\% better than the FDE value of 1.68m by Trajectron++. Furthermore,  Grouptron achieves an FDE of 0.97m on the UNIV dataset. This is 9.3\% reduction in FDE error when compared with the FDE value of 1.07m by Trajectron++ on the same dataset.

Moreover, we compare Grouptron's performance with Trajectron++'s in dense crowds with the UNIV-N datasets in Table \ref{table2}. Overall, Grouptron outperforms Trajectron++ on all the UNIV-N test sets by enormous margins. In particular, Grouptron achieves an FDE of 1.04m and ADE of 0.40m on the UNIV-45 test set, which contains all timesteps from the original UNIV test set that have at least 45 pedestrians in the scene at the same time. This is 16.1\% in FDE improvement when compared with the FDE value of Trajectron++ and 13.0\% in ADE improvement when compared with the ADE value of Trajectron++ on the same test set. 

Furthermore, we notice that the state-of-the-art method, Trajectron++, performs substantially worse as the number of pedestrians in the scene increases. Specifically, Trajectron++'s FDE increases from 1.07m to 1.25m as the minimum number of pedestrians in the scene increases from 1 to 50. In contrast, Grouptron's FDE remains relatively stable as the number of pedestrians increases. This shows that Grouptron performs much better and is more robust in densely populated scenarios. 

\subsection{Qualitative Analysis}
\begin{figure}[t]
    \vspace{3mm}
      \centering
      \includegraphics[scale=0.33]{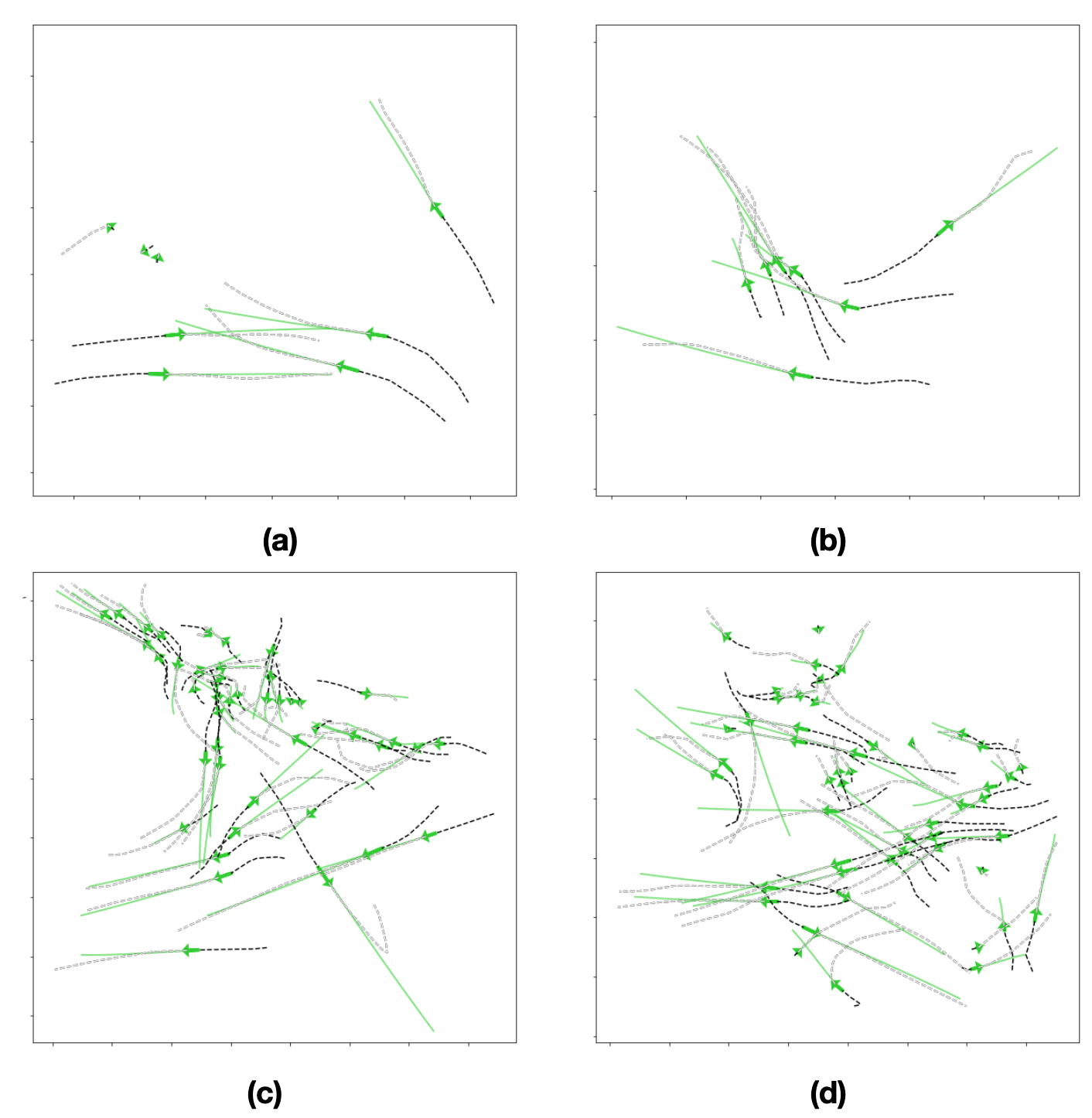}
      \caption{Examples of Grouptron's predictions on the UNIV dataset. The predictions are the most likely trajectory predictions of the model. Green arrows indicate pedestrians' current positions and directions.  Black dashed lines indicate trajectory histories. Grey dashed lines indicate ground truth future trajectories. Green lines are Grouptron's predicted future trajectories. }
      \label{qualatative}
   \vspace{-3mm}
   \end{figure}
   \begin{figure}[t]
      \centering
      \includegraphics[scale=0.30]{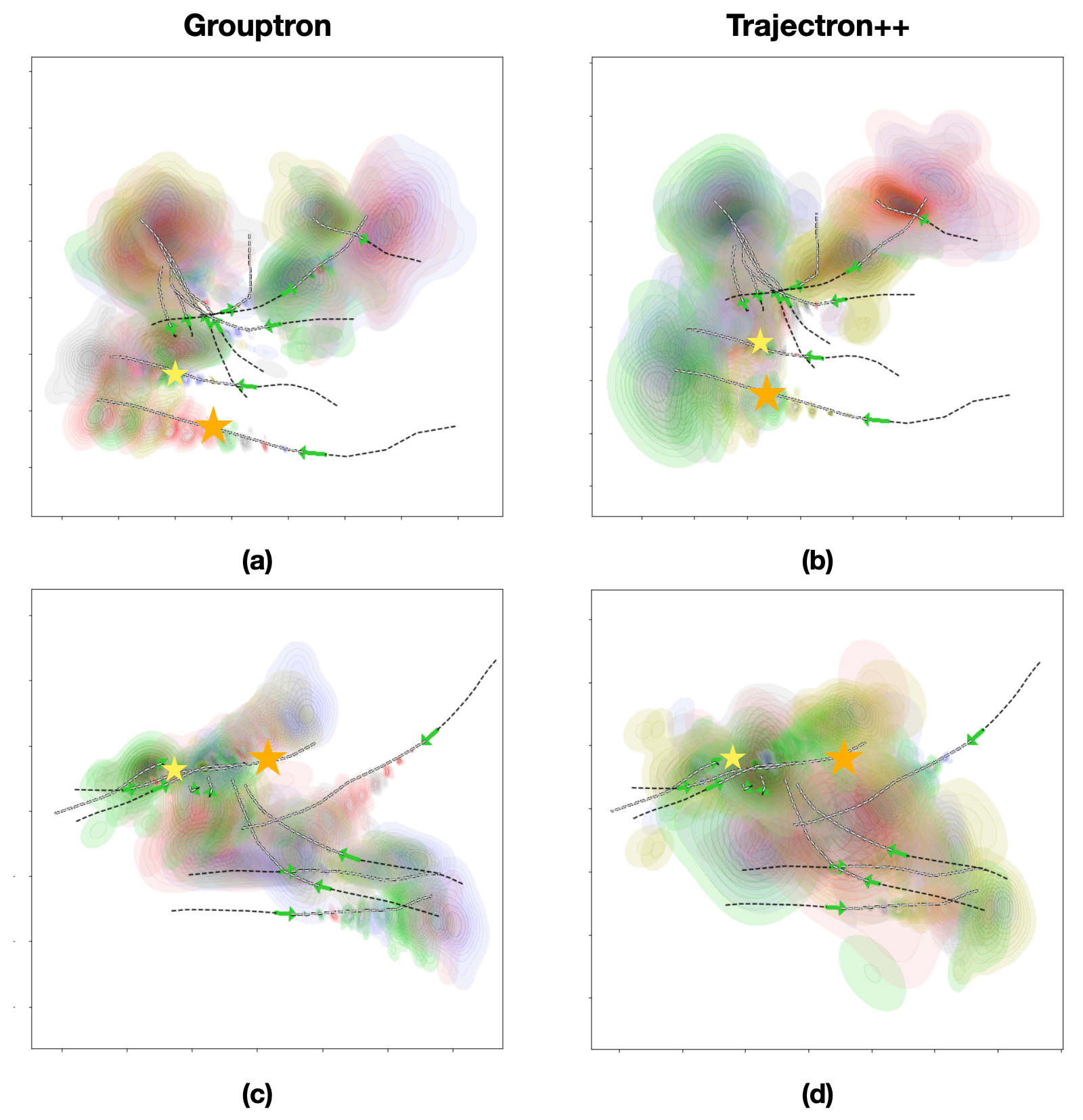}
      \caption{Comparisons of Grouptron's and Trajectron++'s distributions of 20 most likely future trajectories on examples of the UNIV dataset. Rows indicate different examples and columns represent different methods. Orange stars indicate pedestrians of interest. Yellow stars indicate their companions in the same pedestrian groups. Green arrows indicate pedestrians' current positions and directions.  Black dashed lines indicate trajectory histories. Grey dashed lines indicate ground truth future trajectories. The comparisons show that Grouptron is able to produce predictions with higher quality and with better confidence levels.}
      \label{qualatative2}
   \vspace{-3mm}
   \end{figure}
Fig. \ref{qualatative} shows Grouptron's most likely predictions for some examples of the UNIV dataset. Fig. \ref{qualatative}a shows two pedestrian groups crossing paths. We can see that Grouptron's predictions are consistent with the groups. Furthermore, it accurately predicts when and where the two groups' trajectories intersect. Fig. \ref{qualatative}b shows a case where pedestrians are forming groups and merging paths. Grouptron again successfully predicts the formation of this group. Fig. \ref{qualatative}c and Fig. \ref{qualatative}d show Grouptron's performance in densely populated scenes with more than 40 pedestrians. Even in these extremely challenging scenarios for state-of-the-art methods, Grouptron still produces predictions of high quality and the predictions are consistent with pedestrian groups. Furthermore, we can see that even when pedestrian groups are crossing paths or influencing each other, Grouptron successfully predicts these highly dynamic and complex scenarios.

 In Fig. \ref{qualatative2}, we compare Grouptron's distributions of 20 most likely predictions with those of Trajectron++'s. Comparing Fig. \ref{qualatative2}a with \ref{qualatative2}b and \ref{qualatative2}c with \ref{qualatative2}d shows that Grouptron's predictions for the pedestrians of interest reflect the interactions within pedestrian groups more accurately. Furthermore, Grouptron's prediction distributions have much smaller ranges, indicating that it is much more confident with prediction outcomes.

\section{Conclusions}

In this paper, we present Grouptron, a multi-scale graph neural network for pedestrian trajectory forecasting. The spatio-temporal graphs model complex and highly dynamic pedestrian interactions at three scales: the individual level, the group level, and the scene level. Grouptron detects pedestrian groups using agglomerative clustering based on Hausdorff distances, uses GCNs and LSTM-based networks to encode information at different scales, and combines the embeddings across scales. 
With these novel designs, Grouptron achieves high performance even in situations that are tough for state-of-the-art methods. Through extensive experiments, we show that Grouptron outperforms state-of-art methods on ETH and UCY datasets, which are standard benchmarks in the field. Furthermore, through qualitative comparisons and experiments run on densely populated UNIV-N test sets, we show that Grouptron outperforms state-of-art methods by even larger margins and is robust to the change in the number of pedestrians in the scene.

\bibliographystyle{IEEEtran}
\bibliography{reference.bib} 

\addtolength{\textheight}{-12cm}

\end{document}